\documentclass[a4paper]{article}

\usepackage{ISCSLP2022}
\usepackage{CJKutf8}
\usepackage{color}
\usepackage{makecell}
\usepackage{enumerate}
\usepackage{cite}

\title{Improving Rare Words Recognition through Homophone Extension and Unified Writing for Low-resource Cantonese Speech Recognition}

\name{HoLam Chung$^{1,3,^\dagger}$\thanks{$^\dagger$Work done during his internship at CPII.}, Junan Li$^{1,2}$, Pengfei Liu$^{1,^\ast}$\thanks{$^\ast$Corresponding author.}, Wai-Kim Leung$^{1}$, Xixin Wu$^{1,2}$, Helen Meng$^{1,2}$}
\address{
  $^1$Centre for Perceptual and Interactive Intelligence (CPII), Hong Kong \\
  $^2$The Chinese University of Hong Kong, Hong Kong \\
  $^3$Graduate Institute of Electrical Engineering, National Taiwan University, Taiwan
}
\email{holam.chung@protonmail.com, \{jali,pfliu,williamleung,wuxx,hmmeng\}@cpii.hk}

\begin{document}

\maketitle
\begin{abstract}
Homophone characters are common in tonal syllable-based languages, such as Mandarin and Cantonese. The data-intensive end-to-end Automatic Speech Recognition (ASR) systems are more likely to mis-recognize homophone characters and rare words under low-resource settings. For the problem of low-resource Cantonese speech recognition, this paper presents a novel homophone extension method to integrate human knowledge of the homophone lexicon into the beam search decoding process with language model re-scoring.
Besides, we propose an automatic unified writing method to merge the variants of Cantonese characters and standardize speech annotation guidelines, which enables more efficient utilization of labeled utterances by providing more samples for the merged characters.
We empirically show that both homophone extension and unified writing improve the recognition performance significantly on both in-domain and out-of-domain test sets, with an absolute Character Error Rate (CER) decrease of around 5\% and 18\%.
\end{abstract}

\section{Introduction}

End-to-end deep learning models \cite{graves2014towards,amodei2016deep,watanabe2017hybrid,battenberg2017exploring,li2020towards} have become more and more popular in developing speech recognition systems because of simpler architecture and competitive performance compared with classical modular-based or hybrid systems. However, these models are typically data-intensive and perform worse on low-resource settings.
With the recent progress of self-supervised learning in speech and language processing, it has become a common practice to fine-tune a pre-trained model for low-resource speech recognition. However, it is still very challenging to fine-tune a model with very limited labeled data and especially difficult to correctly recognize rare or unseen words (e.g., entity names) in the training set.
The performance becomes even worse when the test data is very different from the training dataset, known as out-of-domain. For example, we observe a significant CER increase from 13\% in an in-domain test set to 50\% in an out-of-domain test set where many characters or words are rare or unseen in the training set.


\begin{CJK*}{UTF8}{bsmi}
\begin{table}[htb]
\centering
\caption{Examples of Cantonese homophone characters and the variants of characters with same Jyutping and semantic.}
\vspace{-0.5em}
\resizebox{0.66\linewidth}{!}{%
\begin{tabular}{cl}
\hline
\textbf{Jyutping} & \textbf{Homophone Characters} \\ \hline
zo2 & 左, 阻, 俎, 柤, 詛, 座 \\
sai3 & 世, 細, 勢, 婿, 貰, 些, 僿, 埶, 楴 \\
wong4 & 王, 黃, 皇, 簧, 煌, 蝗, 惶, 磺, 凰 \\ \hline
\textbf{Jyutping} & \textbf{Variants in Writing} \\ \hline
zoeng3 & 帳, 賬 \\
lei5 & 裏, 裡 \\
zeng6 & 淨, 凈 \\ \hline
\end{tabular}%
}
\label{tbl:example}
\end{table}
\end{CJK*}



In this paper, we focus on Cantonese speech recognition and aim to improve the recognition performance of words that have rare occurrences or are unseen in the training set. 
Compared with English, Cantonese is a tonal language with a lot of homophone characters and words, which bring a further challenge for an ASR model trained with limited data. As illustrated in Table~\ref{tbl:example}, different Cantonese characters share identical pronunciations (i.e., homophone characters) denoted using Jyutping, which is a Cantonese romanization system developed by the Linguistic Society of Hong Kong. Besides, a character may have several variants (i.e., same pronunciation, same meaning but different ways of writing).

For the low-resource setting of Cantonese speech recognition, homophone characters and writing variants pose the two challenging issues: (1) a character may be mis-recognized as its homophone characters which may have higher frequency in the limited training set; (2) the variants of a character may lead to inconsistent transcripts by different annotators or from different datasets, which reduce the number of training samples for the character and introduce unnecessary complexity in the decoding process. 
The two issues may worsen when the trained model is evaluated on an out-of-domain dataset, which may have more rare or unseen characters or words than the training set.

As a general definition, we regard the two issues as the problem of rare words recognition, and propose the two methods, namely homophone extension (HE) and unified writing (UW), to improve the recognition performance.
Regarding homophone extension, we exploit a Cantonese lexicon to identify the homophone characters (potentially rare in the training set) and inject them into the beam search process, followed by a language model (LM) re-scoring step. Therefore, those words that are rare or unseen during training can also be recognized with a certain probability during inference.
For the approach of unified writing, we merge the variants of Cantonese characters and unify them with the most frequent one to avoid unnecessary complexities caused by the variants. This step also increase the training samples of a particular character.

We evaluate the aforementioned two methods on three publicly available Cantonese datasets and confirm that both methods improve the recognition performance, particularly on an out-of-domain test set with more rare words, on which the CER was reduced from 50.58\% to 31.87\%.

\section{Related work}

The recent trend of using pre-trained self-supervised learning language models have proven their strong ability to handle different downstream tasks. 
As a result, many works in the literature have begun to explore transfer learning of pre-trained models in speech processing.
For example, \cite{baevski_wav2vec_2020} proposes Wav2Vec2.0, which applies a speech pre-trained technique that allows the model to learn powerful representations from speech audio alone. Along this research direction, \cite{babu2021xls} utilize the Wav2Vec2.0 model in a multilingual setting and propose the XLS-R model pre-trained on 128 different languages. The downstream ASR task fine-tuned on the XLS-R model outperforms other pre-trained speech models, especially in low-resource languages.
To improve the amount and quality of Cantonese speech data \cite{yu_automatic_2022} has proposed a new corpus called Multi-Domain Cantonese Corpus (MDCC), which is collected from audiobooks from Hong Kong, trying to cover as many domains as possible. They also address the problem of Cantonese character variants that share the same meaning and pronunciation by manually unifying their writing.

Some recent works have pointed out the challenges of recognizing rare words by just fine-tuning the pre-trained speech models. For example, \cite{yang2021multi} presents a multitask training of a language model that utilizes the intent and slots as additional targets and uses the language model in a second-pass re-scoring to improve the recognition performance of rare words (e.g., slots) for the specific application of an interactive voice assistant. Moreover, \cite{ravi2020improving} utilizes the unigram shallow fusion method to apply a fixed reward when a rare word is encountered during decoding. However, this method does not work if the rare words do not appear during decoding. \cite{sainath2021efficient,wang2022improving} explore the idea of using a hybrid autoregressive transducer with a neural language model to re-rank the output of the acoustic model for a better rare word quality. Besides using LM, \cite{zheng_using_2021} proposed using speech synthesis techniques to synthesize audios of the rare or out-of-domain words and use the synthesized audios to fine-tune the ASR model for better recognition accuracy. Lastly, \cite{bruguier2019phoebe} proposes an approach named Phoebe, which extends the contextualized Listen, Attend, and Spell (CLAS) model by injecting pronunciations into the context module and provides a 16\% relative word-error-rate reduction over CLAS. Phoebe has access to both the textual form of the bias phrases and their corresponding pronunciations and thus improves the recognition performance of rare or unseen words.


\section{Methodology}
It is a common practice to fine-tune a pre-trained speech model for ASR development. 
However, the rare or unseen words problem due to lack of labeled data still can not be well addressed by simply fine-tuning the XLS-R model with the limited data.
To alleviate this problem, we propose a homophone extension (HE) approach where human knowledge of rare words based on a Cantonese lexicon is injected to the beam search process, followed by a language model re-scoring step. In this way, those words that are rare or unseen during training can also be recognized with a certain probability during inference.
Furthermore, we also merge the variants of Cantonese characters to the most frequently used one to avoid unnecessary complexities caused by the variants and increase the training samples of a particular character. This process is called automatic unified writing (UW), which also reduces CER in our experiments.

\subsection{Baseline Model}

We choose to fine-tune the pre-trained model named XLS-R \cite{babu2021xls} as our baseline model, which adopts a linear layer on top of the pre-trained model trained with the Connectionist Temporal Classification (CTC) loss.
The decoding process is conducted by a beam search algorithm, which keeps the top B candidates for each decoding step and prunes the unlikely paths based on the tree of the output sequence.
An n-gram LM shallow fusion is also integrated in the beam search decoding process to produce more semantically related results. However, the improvement is not significant in terms of recognizing rare words and unseen words. The top B candidates may not contain the target words due to lack of training data that covers these words. 

\subsection{Homophone Extension}
We propose a Homophone Extension (HE) method to improve the recognition of rare or unseen words in the low-resource training set. 
The main idea is to inject a homophone lexicon into the beam search process with a specify initial score to allow the homophone rare words to appear in the search space with re-estimate score.
To determine the homophone characters of a given character, we adopt a publicly available lexicon for the Jyutping\cite{linguistic1997hong} input method, which maps a Jyutping code (e.g., wong2) and its characters together, known as homophones. 

\begin{figure}[htb]
  \centering
  \includegraphics[width=0.46\textwidth]{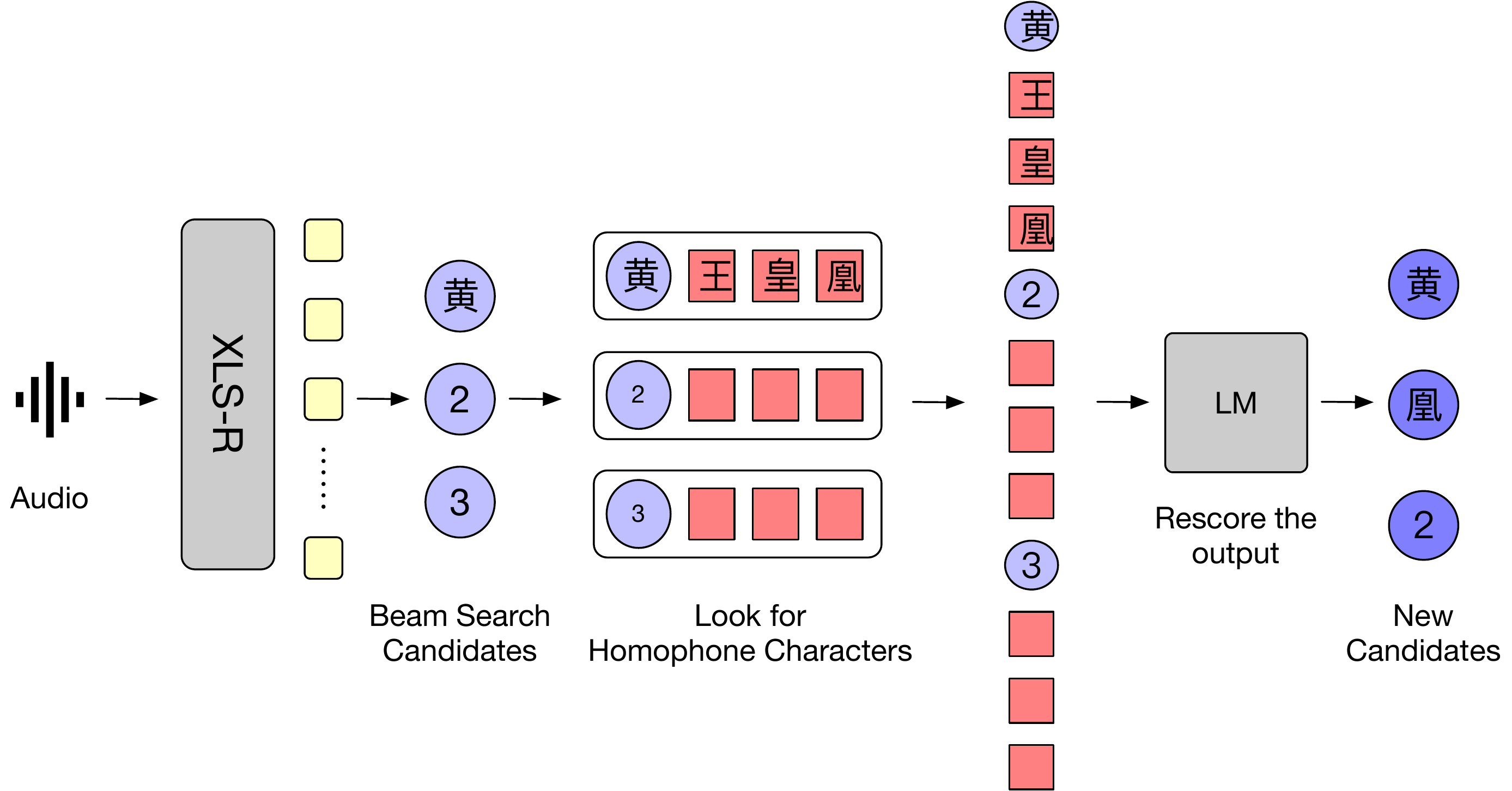}
  \vspace{-0.5em}
  \caption{The homophone extension method injected in the beam search process and language model re-scoring.}
  \label{fig:homophone-extension}
\end{figure}

As illustrated in Figure~\ref{fig:homophone-extension}, an audio utterance is fed into a fine-tuned XLS-R model which outputs a probability distribution for each span of frames and the top B (e.g., B=3 in the figure) candidate characters are kept after pruning in the beam search algorithm.
For each candidate character, we extend the search space by adding their homophone characters. Then, a language model is adopted to re-score all the extended candidates and output the final new candidates.
Remarkably, the proposed homophone extension method is comparable with any language model re-scoring approach for speech recognition.
Considering those homophone characters may be very low and not able to appear in the top-rank result, we further adjust the output probability of each homophone to the corresponding source char, which has a much higher probability than the homophone character's output.


Formally, for a given character $c$, let $a_p(c)$ be the acoustic probability from the CTC model, and $H_c = \{h_i^c\}$ be the homophone character set of $c$ where $i \in \{1,\ldots,M\}$. A character may have several different pronunciations, called polyphone, and we denote $N(h_i^c)$ as the number of pronunciations for the character $h_i^c$, we get $h_i^c$ from Jyutping that sharing the same code with tone. And $q(h_i^c)$ as the acoustic probability from the CTC model of that polyphone.
We also introduce a discount factor $(1-\log_{10} N(h_i^c))$ to $q(h_i^c)$ to lower the probability of a polyphone with more pronunciations, as well as a weight factor $\gamma$ to balance the probabilities of $a_p(c)$ and $q(h_i^c)$. 
Besides, we apply a $\max$ operation to ensure the updated probability $p(h_i^c)$ is not lower than the original probability $a_p(c)$, as illustrated in Equation (1):
\begin{multline}
    p(h_i^c) = \max \{a_p(c), (1-\gamma) \cdot a_p(c) + \\
    \gamma \cdot q(h_i^c) \cdot (1-\log_{10} N(h_i^c)) \}
\end{multline}


\subsection{Unified Writing}
There is a lack of rigorous Cantonese transcribing standards, whereas many characters share the same pronunciation and meaning in similar writing or typing code. 
\begin{CJK*}{UTF8}{bsmi}
Moreover, the language styles between different datasets are also different, the annotation variation is increased when different datasets are combined. For example,  裏面 (inside) and 裡面 (inside) are two words that share the same meaning, only ``裏面" will appear in the MDCC\cite{yu2022automatic} dataset, but ``裏面" and ``裡面" will both appear in the Common Voice Dataset.
\end{CJK*}
There are two critical issues for this problem - the first is to find as many as possible variants of a character, and the second is to check whether replacing the character with its variants will affect the original meaning or not. 

Rather than relying on manual replacement, which requires linguistic expert knowledge and costly checking, we propose an automatic approach called Unified Writing (UW) to handle annotation variation across datasets. This approach consists of the first step of \textit{unified pair discovery} and the second step of \textit{replacement with a rewriting semantic checker}. We assume that the annotation variations come from the typing system of annotators. There are many input methods in Cantonese like Jyutping, Cangjie, 4Corner, etc. These typing methods can be categorized as pronunciation encoding or glyph encoding. The similar character in pronunciation or glyph, the typing codes based on pronunciation or glyph will also be similar.



Let $C_{0..L}$ be all the characters in the lexicon, where $L$ is the total number of entries in the lexicon. The unified pair discovery will iterate all the character combinations of $L \cdot (L-1)$ in total. During the iteration process, we calculated the pronunciation encoding similarity from Jyutping with normalized edit distance, and then we filtered out the character combinations with a Jyutping edit distance $d > 0$, which means they have different pronunciations.
For glyph encoding similarity, we generate the typing code using the typing methods respectively using Changjei5, Simplex5, BSM, CKC, QCode, G6Code, Stroke5, Boshiamy, DaYi4, 4Corner5, an then filter all the results with a normalized edit distance of $d > 0.25$.


After the two filtering iterations, we can get a list of character pairs that have the same pronunciation and similar glyphs. To make sure the usage of character is similar, we further use fastText \cite{joulin2017bag,bojanowski2017enriching} to filter the semantic dissimilar pair with a cosine similarity value $> 0.5$. 
The character pair with similarity value close to 1 means the two characters are more likely to appear in the same context. 
After the filtering, the remaining character pairs are the output of the unified pair discovery step, and are potentially replaceable since they have met with all the character similarity metrics.

The next step is to conduct the actual replacement based on frequency if it passes a rewriting semantic checker. A character with high frequency means it is more common in the target application scenarios. We change the lower-frequency characters to the higher-frequency ones. After the replacement, we also need to justify the meaning after the unified writing. We adopt the BERT score \cite{zhang2019bertscore} to further evaluate the semantic similarity between the original transcript and the replaced one, where the higher BERT score indicates the higher similarity between two transcripts. We filter out the results where the similarity scores are less than $0.9$.




\section{Experiments}

\subsection{Experimental Datasets}
We evaluate the proposed methods on three publicly available datasets, namely MDCC \cite{yu2022automatic}, Common-Voice\cite{commonvoice2020}, and Yueqie\footnote{http://www.livac.org/yueqie/}.
Table~\ref{tbl:datasets} shows the number of utterances in each dataset, where we use the default split of MDCC, and the Cantonese subset of Common-Voice collected by Mozilla\footnote{https://huggingface.co/datasets/mozilla-foundation/common\_voice\_9\_0}. We use all the data of Yueqie with 171 utterances in total, for evaluating the proposed methods on the out-of-domain setting.
In the Yueqie dataset, there are in total 171 4-character words. 25 of them appear only once in the training set, and 4 of 171 show up twice.

\begin{table}[htb]
\centering
\caption{The number of utterances in each experimental dataset.}
\vspace{-0.5em}
\resizebox{0.7\linewidth}{!}{%
\begin{tabular}{|r|c|c|c|}
\hline
            & Train & Dev  & Test  \\ \hline
MDCC\cite{yu2022automatic}        & 65120 & 5663 & 12492 \\ \hline
Common-Voice\cite{commonvoice2020} & 8392  & 5578 & 5578  \\ \hline
Yueqie    & -     & -    & 171   \\ \hline
\end{tabular}}
\label{tbl:datasets}
\end{table}

\subsection{Character Tokenizer}
We build our ASR system using a large character tokenizer to cover as many Cantonese characters as possible and include the rare characters that may not appear in our dataset. Otherwise, those unseen characters will be recognized as unknown during inference, and can not support model adaptation to a new domain with new words. Besides, we need a larger tokenizer to support the homophone extension method, which may introduce some rare or unseen words in the decoding process.
We collect all the characters from the open source project that provides a cin table for the input method, it is a query table that links the typing code with the character\footnote{https://github.com/chinese-opendesktop/cin-tables}.
We use the cin table in Chinese (Hong Kong) and get all the unique characters. It will cover all the possible characters that the annotator may use, and the final tokenizer consists of 32,693 characters. 

\subsection{Implementation Details}
Follow the fine-tuning setting from XLS-R\cite{babu2021xls}, we conduct experiments using \texttt{xls-r-300m} as the backbone model and fine-tune it using the CTC loss to build our ASR. The model is trained on 6 GPUs with a batch size of 17 and a gradient accumulation of 20. The learning rate is 2e-3 and will linearly decay during the whole training process. We train our model with 150 epochs and choose the best checkpoint based on the development set for further testing.
For language model integration, we train a 5-gram KenLM model using the Cantonese Wikipedia dataset\footnote{https://dumps.wikimedia.org/zh\_yuewiki}.
The hyper-parameters for the beam-based decoding are set as $\alpha=0.45$, $\beta=1.55$ and the beam size as 20, where $\alpha$ is the weight for language model shallow fusion and $\beta$ is the weight for score adjustment based on the length of the decoded sequence\cite{kannan2018analysis}.
Besides, the $\gamma$ for homophone extension in Equation (1) is set as 0.5.


\subsection{Experimental Results}
Table~\ref{tbl:testset-cer} presents the performance comparisons among different test sets in terms of CER (\%), where Yueqie is a different out-of-domain dataset on which the model shows very high CER.

\vspace{-0.5em}
\begin{table}[htb]
\centering
\caption{Performance comparisons among different test sets.}
\vspace{-0.5em}
\resizebox{0.6\linewidth}{!}{%
\begin{tabular}{|r|r|}
\hline
\textbf{Test Set}    & \textbf{CER(\%)}  \\ \hline \hline
MDCC        & 9.32 \\ \hline
Common-Voice & 24.93 \\ \hline
MDCC + Common-Voice     & 13.88 \\ \hline \hline
Yueqie      & 50.58 \\ \hline
\end{tabular}}
\label{tbl:testset-cer}
\end{table}

Table~\ref{tbl:method-cer} shows the performance comparisons among different methods on the in-domain dataset (i.e., MDCC + Common-Voice, and the out-of-domain dataset (i.e., Yueqie) respectively. It can be seen that both HE and UW are effective in reducing the CER, particularly on the Yueqie dataset. Note that although we adopt UW on the training set in some settings, we do not apply the UW process on the testing set for all the experiments to make the results comparable.
Furthermore, we evaluate the methods on the testing data with UW, which automatically changes some of transcripts in MDCC and Common-Voice, but keeps the Yueqie dataset unchanged due to no matching characters. We observe that all the method combinations reduce the CER up to 7.73, thanks to merged training samples and testing transcripts by unified writing. 
\begin{table}[htb]
\centering
\caption{Performance comparisons among different methods on MDCC + Common-Voice (in-domain) and Yueqie (out-of-domain).}
\vspace{-0.6em}
\resizebox{0.9\linewidth}{!}{%
\begin{tabular}{|l|r|}
\hline
\textbf{Dataset \& Method}  & \textbf{CER (\%)}    \\ \hline
\multicolumn{2}{|l|}{\textit{MDCC + Common-Voice without UW on Test Set}} \\ \hline \hline
baseline                                    & 13.88  \\ \hline
baseline with beamsearch \& lm              & 10.66  \\ \hline
baseline with beamsearch \& lm \& HE        & 10.28  \\ \hline
baseline with beamsearch \& lm \& UW       & 8.61   \\ \hline
baseline with beamsearch \& lm \& HE \& UW & \textbf{8.07}   \\ \hline \hline
\multicolumn{2}{|l|}{\textit{MDCC + Common-Voice with UW on Test Set}} \\ \hline \hline
baseline with beamsearch \& lm              & 11.17  \\ \hline
baseline with beamsearch \& lm \& HE        & 10.82  \\ \hline
baseline with beamsearch \& lm \& UW       & 8.05   \\ \hline
baseline with beamsearch \& lm \& HE \& UW & \textbf{7.73}   \\ \hline \hline
\multicolumn{2}{|l|}{\textit{Yueqie with / without UW on Test Set}} \\ \hline \hline
baseline                                    & 50.58  \\ \hline
baseline with beamsearch \& lm              & 39.33  \\ \hline
baseline with beamsearch \& lm \& HE        & 35.82  \\ \hline
baseline with beamsearch \& lm \& UW       & 33.78  \\ \hline
baseline with beamsearch \& lm \& HE \& UW & \textbf{31.87}  \\ \hline 
\end{tabular}}
\label{tbl:method-cer}
\end{table}

\subsection{Case Studies}

To better understand how HE improves the recognition of rare words, we analyze some examples from the Yueqie dataset and compare the results with and without HE. As shown in Table~\ref{tab:HE case study}, the first four utterances are correctly recognized with HE, although the last two utterances are mis-recognized even from the first character. In fact, homophone extension has improved the recognition performance for 20 of 171 Cantonese 4-character idioms, and failed to improve the other 6 idioms (e.g., 5th and 6th in Table~\ref{tab:HE case study}) where the mis-recognition of the first character will very likely cause the subsequent language model re-scoring to fail. Note that the empty red blocks imply the corresponding characters are not recognized.
\begin{table}[htb]
\caption{Prediction examples of with and without HE.}
\label{tab:HE case study}
\vspace{-0.6em}
\centering
\includegraphics[width=0.46\textwidth]{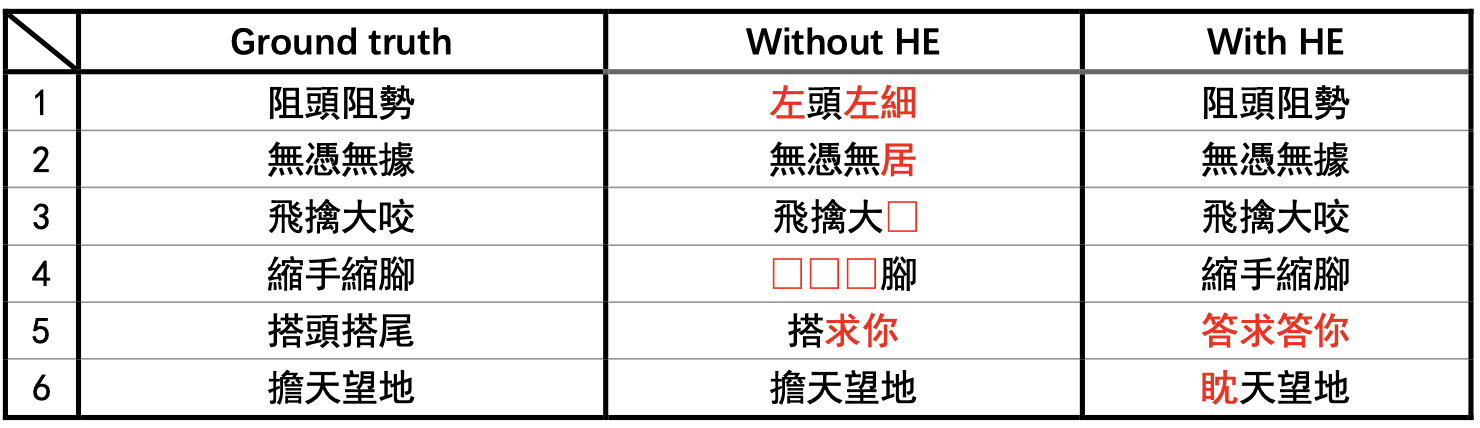}
\end{table}

Table~\ref{tab:UW case study} presents some prediction examples with and without using unified writing (UW). The characters highlighted in blue have their corresponding variants. As shown in the second column of Table~\ref{tab:UW case study}, these characters would be mis-recognized by the model (highlighted in red) without UW. However, if we unified these words into their variants, which are more frequently seen in the corpus, the model manages to recognize them as their corresponding variants (highlighted in green). 
Given that the ground truth transcription still uses the character variants for labeling, the CER is not reduced for the examples in Table~\ref{tab:UW case study}. However, the predictions with UW are actually correct and the variants are more commonly used in Cantonese writing.
\vspace{-0.5em}
\begin{table}[htb]
\caption{Prediction examples with and without UW.}
\label{tab:UW case study}
\vspace{-0.6em}
\centering
\includegraphics[width=0.46\textwidth]{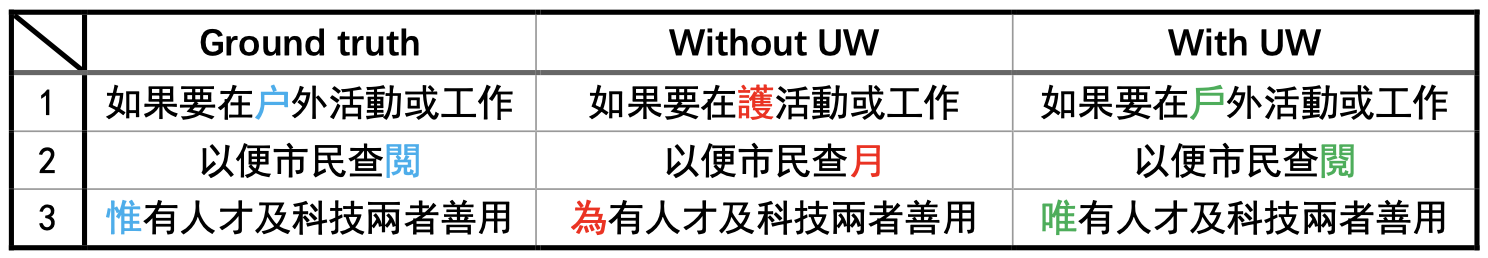}
\end{table}

\section{Conclusions}

This paper presents a novel homophone extension method and an automatic unified writing approach to improve the recognition of rare or unseen characters in the training dataset for the low-resource setting of Cantonese speech recognition. With the help of a Cantonese homophone lexicon, this method enables a rare character to appear in the beam search space if one of its homophone characters is encountered in the decoding process. Through using the unified writing approach, we utilize the labeled utterances more efficiently and improve the recognition of Cantonese character variants. Experimental results show that the two proposed methods improve the recognition performance significantly, particularly on the out-of-domain dataset with a dramatic CER drop from 50\% to 31\%.
The major contributions of this paper, focusing on the problem of low-resource Cantonese speech recognition, are summarized as follows:
\begin{enumerate}[(1)]
\setlength{\itemsep}{0pt}
    \item Propose a homophone extension method to integrate the human knowledge of a homophone lexicon into the beam search process and improve recognition of rare words through their homophone characters;
    \item Introduce an automatic unified writing approach to merge the variants of a Cantonese character and lead to performance gains due to more training samples for the character and reduced decoding space;
    \item Show empirically that the proposed two methods improve the recognition performance significantly on both in-domain and out-of-domain datasets.
\end{enumerate}


\section{Acknowledgements}
This work is partially supported by the Centre for Perceptual and Interactive Intelligence (CPII) Ltd., a CUHK-led under the InnoHK scheme of Innovation and Technology Commission; and in part by the HKSAR RGC GRF (Ref No. 14207619). We thank Jeff Chen from the National Taiwan University for his helpful advice and comments on this paper.

\bibliographystyle{IEEEtran}
\bibliography{mybib}

\end{document}